# Topic Modelling on Consumer Financial Protection Bureau Data: An Approach Using BERT Based Embeddings

Vasudeva Raju S.
UpGrad Education,
Mumbai, India

Bharath Kumar Bolla
Salesforce India
Hyderabad, India

Deepak Kumar Nayak
Brane Enterprises
Hyderabad, India

Jyothsna Kh
Global Logic
Hyderabad, India

*Abstract*—Customers' reviews and comments are important for businesses to understand users' sentiment about the products and services. However, this data needs to be analyzed to assess the sentiment associated with topics/aspects to provide efficient customer assistance. LDA and LSA fail to capture the semantic relationship and are not specific to any domain. In this study, we evaluate BERTopic, a novel method that generates topics using sentence embeddings on Consumer Financial Protection Bureau (CFPB) data. Our work shows that BERTopic is flexible and yet provides meaningful and diverse topics compared to LDA and LSA. Furthermore, domain-specific pre-trained embeddings (FinBERT) yield even better topics. We evaluated the topics on coherence score (c_v) and UMass.

## I. Introduction

Customer complaints and reviews are important topics in every industry. Complaints and reviews benefit businesses indirectly by allowing them to hear from their customers about their products and services. Those comments and reviews might be positive or negative. Customers can file complaints in two ways: offline and online. Customers must manually complete a feedback form in the offline option, whereas the online mode is open-ended. Online complaints can be made via video, audio, image, or text. As every business is going digital, it's getting increasingly difficult for companies to manage online complaints and reviews. Topic modelling can help classify customer complaints, and it allows financial organizations to extract the customer's most pressing issues from their complaint narrative. In addition, Topic modeling can identify the root causes of low consumer satisfaction.

The two most popular and conventional techniques for topic modeling are Latent Semantic Analysis (LSA) and Latent Dirichlet Allocation (LDA). LSA is a deterministic technique, whereas LDA is a statistical model. Latent Semantic Analysis examines the relationship between a collection of documents and their terms. It explores unstructured data for hidden links between phrases and concepts using singular value decomposition (SVD). When performed on large corpora, LSA can achieve significant data compression, but being a linear model, it fails to handle nonlinear dependencies. LDA [4] overcomes the deficiencies of LSA and is still the most popular technique for topic modeling. However, LDA ignores topic correlation since it believes all topics are independent [4], and with large corpora, LDA has issues with sparsity [4, 15].

Research into word embeddings LDA gained momentum during the same period. Word Embeddings, a popular technique, trains text representation in n-dimensional space and then groups words with similar meaning in lower-dimensional vector space [2, 3]. Research in word embeddings progressed a lot to handle large vocabularies and datasets. LDA missed semantic relationships between words in a text corpus and failed to work on large corpora.

Recent work combining the techniques of LDA and Word embeddings brought novel ways of utilizing the strengths of LDA and word embeddings, as evidenced by Lda2vec [12]. Embedded Topic Model (ETM) is a framework that combines topic models and word embedding models into a single framework [9].

In this work, we compared and contrasted the topics generated by conventional techniques (LDA & LSA) and an embedding-based novel model called as BERTopic [11] on the Consumer Financial Protection Bureau (CFPB) data. The Consumer Financial Protection Bureau (CFPB) educates and enables individuals to make sound financial decisions, and it enforces federal consumer finance laws by regulating financial products and services for consumers [19]. Our contribution in this work can be summarized as below:

1. Compare and contrast LDA/LSA vs BERTopic based embedding models
2. Evaluate different flavors of BERT models such as BERT, RoBERTa and DistilBERT
3. Evaluate if domain-based finetuned embeddings such as FinBERT performs better than the rest of the BERT variants

## II. Related Work

Topic modelling is an essential and vital study area in Natural Language Processing (NLP). Although topic models were initially designed for text data, they were subsequently used in other data mining domains such as audio, video, and image annotation [10]. Topic modeling techniques can be broadly classified into probabilistic and graphical [13]. Topic models have progressed from Latent Semantic Analysis (LSA) to Probabilistic Latent Semantic Analysis (PLSA), then to Latent Dirichlet Allocation (LDA), and eventually to deep learning techniques combined with LDA to generate Lda2vec [12].

Every topic model built to date works on the exact basic inference: "each document is a collection of topics, and each topic is a mixture of words." Latent Semantic Analysis (LSA) is one of the primitive topic modelling techniques that has been studied and evaluated extensively [8]. This simple approach produces topics based on word frequency in each text [14]. This method reduces the dimensionality of a sparse document-term matrix using Singular Value Decomposition (SVD). LSA has the advantage of being simple to implement, comprehend, and computationally fast due to its matrices decomposition. The downside of LSA is that the decomposed matrix is thick, difficult to comprehend, and requires many documents to produce satisfactory results. A refined version of the LSA technique was developed by replacing SVD with mixture decomposition derived from a latent class model. This topic modeling variant was defined as Probabilistic Latent Semantic Analysis (PSLA). Compared to conventional LSA, its probabilistic counterpart has a sound statistical foundation and establishes a proper generative model of the information. The advantage of the probabilistic model is that it can be extended and embedded furthermore easily with other complex models.

LDA [4] is a popular topic modelling technique. LDA creates a 'distribution of distributions' of two matrices: 'Topic per document matrix' and 'Words per topic matrix'. LDA evaluates the distribution of all subjects and terms throughout each document. Unlike pLSA, LDA can quickly generalize to new documents. Several variants of LDA exist, including Correlated Topic Modelling (CTM), which improves the accuracy of topic classification [7]. Another type is dynamic topic modelling, which employs topic modelling throughout a period [5]. Hierarchical LDA is a variant of LDA in which a tree of topics is constructed rather than a uniform topic structure [18]. LDA ignores topic correlation since it believes all subjects are independent [4]. Text and social media data, in particular, have such linkages. This limits the LDA's ability to process vast data sets and make predictions for new documents. The document independence assumption is incorporated in the Dirichlet probability [4]. Word order inside a phrase and temporal interactions between texts are important textual or linguistic aspects [6]. The independence assumption is built into Dirichlet distribution algorithms [4]. These models required new techniques to incorporate topic and association analysis. LDA is also subject to "order effects, "or differences in outcomes due to the order of the training data used [1].

Thousands of tweets, status updates, and comments are created every day. The short text is highly sparse due to the minimal distribution of words across documents. Different subject modelling approaches were discussed [17]. Another author [16] compared the author-topic model and the LDA approach on Twitter. [20] conducted a comprehensive review of various applications and approaches on diverse short text datasets. The extracted topics can be labeled and used for classification of documents [21] in English and can be used to classify native languages [22].

The majority of NLP problems requiring text processing have been examined using word embeddings and topic models. In 2003, seminal research articles on topic modelling using LDA and word embeddings were published. Word embeddings gained popularity and were expanded in various ways to accommodate large vocabularies and massive datasets. Topic models cannot capture the critical semantic relationship between words in a text corpus. On the other hand, word embedding models are focused on capturing the semantic relationship in a lower-dimensional vector space.

### III. DATA SOURCING

This study uses a single dataset that contains consumer complaints about financial companies. The dataset was compiled directly from the Consumer Financial Protection Bureau (CFPB). This information is available in CSV format for download and comprises 6,84,653 text-based complaints as of 23rd March 2021. Essential features are mentioned in the table below (Table 1):

TABLE 1. CONSUMER COMPLAINT DATASET DICTIONARY

| Column Name | Data type |
| --- | --- |
| Date received | date |
| Product | string |
| Company | string |
| Consumer complaint narrative | string |
| Complaint ID | numeric |

- Date received – the date the consumer's complaint was received by the CFPB.
- Product – refers to the forms of financial product.
- Company – the company's name
- Problem – short description of consumer complaint
- Consumer complaint narrative – the consumer's explanation of the complaint
- Zip code, State– the consumer's state and zip code

### IV. RESEARCH METHODOLOGY

The research methodology followed in this research is described in the below figure (Figure 1). We followed two different approached to generate topics. First is for conventional Topic Modelling techniques such as LSA & LDA, where we normalized the text using lemmatization and removed stop words. In the second approach, to preserve the semantic meaning of words and also consider correlated topics, we used BERTopic architecture (Figure 1).

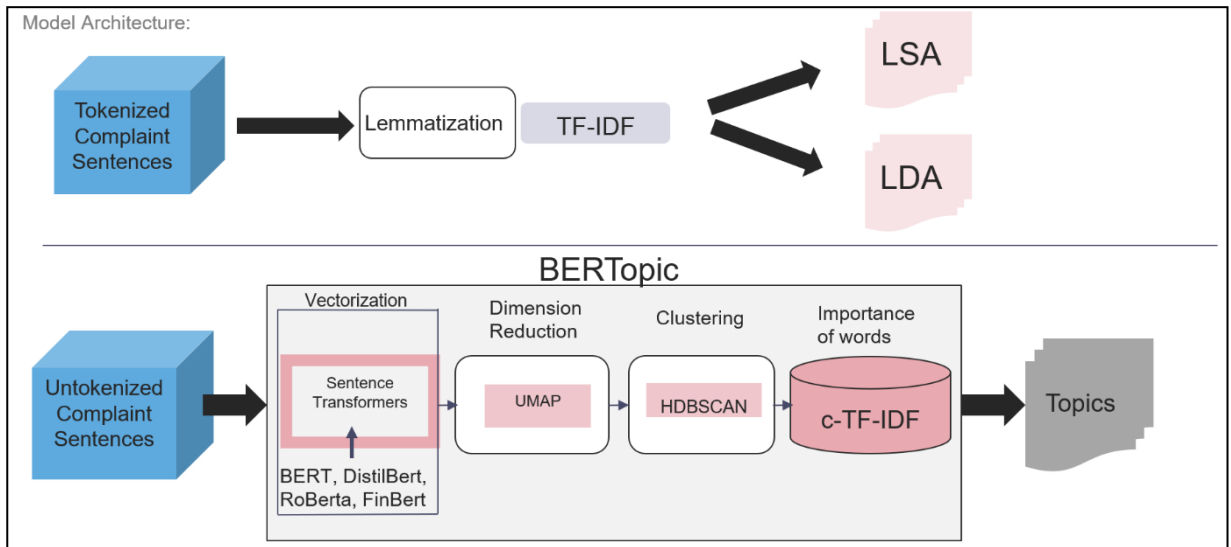

Figure 1. Research Methodology: LSA & LDA (Top); BERTopic (Bottom)

BERTopic has three main algorithmic components: 1) Document Embedding, 2) Document Clustering and, 3) Topic Representation. Sentence Transformers were used to extract document embeddings for the first step of generating document embeddings. BERTopic extracts document level embeddings using the "Para-MiniLM-L6-v2" sentence transformer model. Additionally, we used models such as RoBerta, DistilBERT, and Finbert by employing the Flair package. In the second step, UMAP algorithm is applied to the document embeddings to reduce the dimensions and then HDBSCAN algorithm to cluster the semantically similar documents. Finally, in the third step, we use c-TF-IDF, extract significant words for each cluster on a class-by-class basis (class-based term frequency, inverse document frequency). By computing the frequency of a word in a given document and measuring the word's prevalence across the corpus, TF-IDF enables the comparison of the relevance of terms between documents. Instead, if we treat all documents in a single cluster as a single document and then apply TF-IDF on it, we will obtain relevance scores for individual words within the cluster. The more significant the words inside a cluster, the more representative of that topic they are. As a result, we can acquire topic-specific descriptions. This is extremely useful for inferring meaning from any unsupervised clustering technique's groupings.

## V. RESULTS

The 'C_V' and 'U_Mass' evaluation metrics are used to compare the models. This section also discusses the findings of the LSA, LDA, BERT, FinBERT, DistilBERT, and RoBERTa topic modelling approaches.

### A. LSA Topic model

A form of vector space model i.e., LSA uses the TF-IDF based vectorization that processes text without regard for word order, syntax, or morphology. From the genism model, an LSA model is constructed, and the top five words within each topic are retrieved and analyzed in (Table 2). For better visualization we are representing the topic words as non-stemmed and non-lemmatized words (Table 2, Table 3).

TABLE 2. TOP 5 WORDS OF LSA TOPIC MODELING

| Topic | Topic words |
| --- | --- |
| Service fee | amount, date, pay, card, charge, debt, call, balance, fee, interest |
| Mortgage loans | report, debt, loan, card, call, charge, check, mortgage, would, get |
| Credit Report | debt, date, get. report, claim, inquiry, card, pay, debt, date, get, charge, usd |
| Disputes and Inquiries | report, remove, allege, dispute, debt, inquiry, account, claim, consumer, collect |
| Debt Collection | Consumer, loan, agency, debt, claim, collect, block, report, section, valid |

### B. LDA Topic model

For extracting topics using LDA, we used Gensim package's LDA module which internally uses online variational Bayes algorithm proposed by [4]. Words inside each topic are inspected, and each list is assigned a single topic word to define it. The top ten words in each of the topic representations are listed in (Table 3):

TABLE 3. TOPICS BASED ON MOST WEIGHTED WORDS

| Topic | Topic words |
| --- | --- |
| Debt Collection | debt, number, call, collect, bill, phone, owe, pay, contact, ask |
| Mortgage Loan | mortgage, home, escrow, property, loan, modify, insure, tax, foreclosure, fargo |
| Credit card reporting | card, transact, deposit, check, charge, fund, fee, money, chase, close |
| Auto Loan | Car, loan, dealership, finance, pay, vehicle, repossess, santand, ally |
| Fraud | theft, ident, consumer, report, victim, block, policy, fraudulent, ftc, affidavit |
| Account Reporting | account, report, fraudulent, author, inquiry, remove, equifax, item, plea, open |
| Late fee payment | loan, late, month, pay, make, fee, payment, interest, balance, card, |
| Account Charge-off | acct, balance, request, open, act, oh, cfr, charge, fraudulent, lieu |

## C. BERT

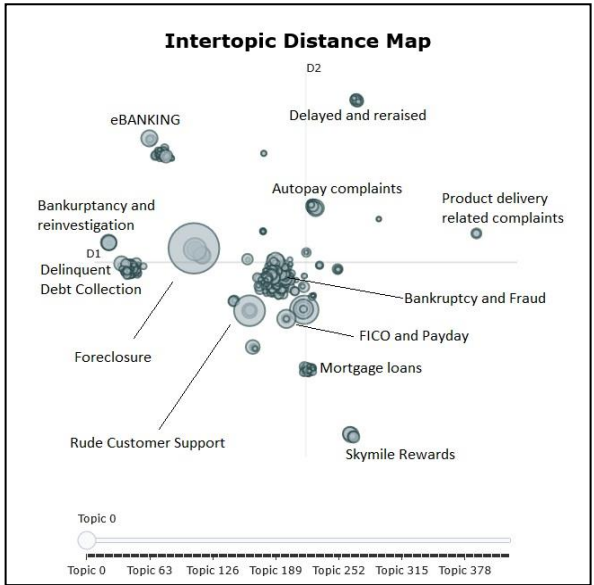

Figure 2. BERT Topic modelling clusters

### TABLE 4. BERT TOPICS

| Topic Name | Major words in the cluster |
|---|---|
| Delayed and reraised Complaints Cluster | Autopay, eLoan, Statement, banker, penalty, alert, tape |
| eBANKING | Payment, Judgement, reinvestigation, Bankruptcy, accounts, trade, major |
| Autopay complaints | Autopay, cardholders, roll, Autotrack |
| Delinquent Debt Collection | Delinquent, limit, Debt, recovery, Collection, discrepancy, major |
| Foreclosure | Foreclosure, Income, value, preserve, equity, deed, specialist |
| Rude Customer Support | dialler, harass, rude, talk, email, recipient, lot |
| Skymile Rewards | Airline, signup, code, advantage, bonus, Skymiles, eligibility |
| Mortgage loans | Mortgage, home, debt, lien, outdate |
| FICO and Payday | Payday, bike, fico, Equifax, experience, inaccuracy, quick |
| Bankruptcy and Fraud | Sadly, Bankruptcy, cheated, scam, game, report, unapproved |
| Product delivery related complaints | Victim, delivery, related, associate, products |

## D. FinBERT

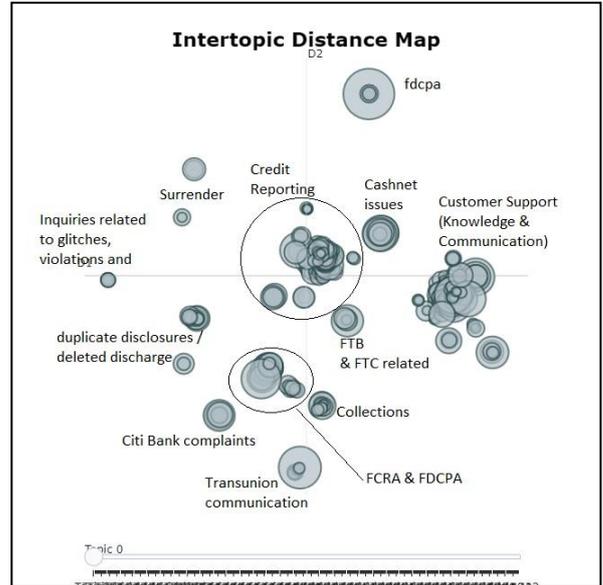

Figure 3. FinBERT Topic modeling clusters

### TABLE 5. FINBERT TOPICS

| Topic Name | Major words in the cluster |
|---|---|
| Credit Reporting | Credit reporting, principal, interest, splitter, overdrawn, overcharge, approximate |
| Glitches, violations related inquiries, and their resolutions | glitch, result, violations, inquiry, percentage, need, resolution |
| Delete discharge / Duplicate disclosures | Damage, delete, diversify, discharge, disclosures |
| Citi Bank complaints | citigold, citibank, citi, eligibility, promotion, promise, consecutive |
| Transunion communication | Transunion, Equifax, statutory, fibula, fibril, fair, private |
| FCRA & FDCPA | fdcpa, fcra, cra, signature, trustor |
| Collections | somebody, information, loan deportment, property, filing, duplicate, collections |
| FTB & FTC | affidavit, ftc, accounts, delete, ftc, debt, delinquency |
| Customer Support (Knowledge & Communication) | harass, voicemail, mislead, criminal, contact, Knowledge, Support |
| Cashnet issues | pennymac, affairs administration, cashnet USA, underwriting, live, payoff |

### E. DistilBERT

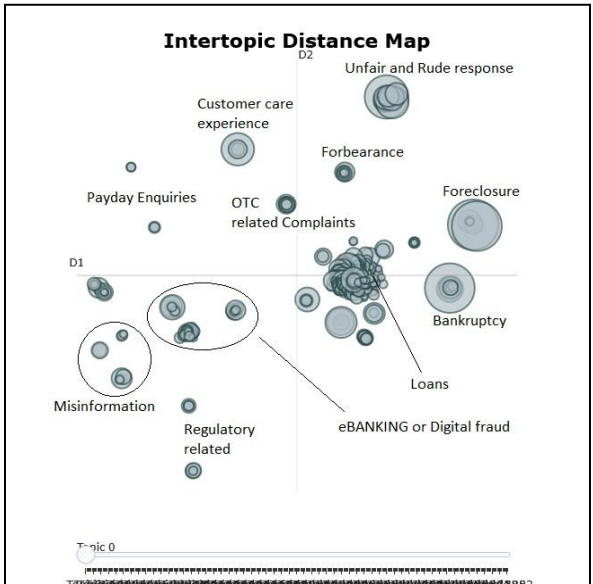

Figure 4. DistilBERT Topic modeling clusters

TABLE 6. DISTILBERT TOPICS

| Topic Name | Major words in the cluster |
|---|---|
| Customer care experience | cancelled in act, chat, consecutive, eligibility, key bank, advantage, response |
| OTC related complaints | cashier, counter, unhappy, teller, frozen, handset, weekly |
| Payday Enquiries | telephone, payday, fraudulent, harass, steady, defer, retail, cheques |
| eBANKING or Digital fraud | glitch, fraudulent, voicemail, cyber, duplicate, telephone |
| Misinformation | Inquiry, Misinformation, negative, share, satisfy, past, fair |
| Regulatory related | Regulatory, surrender, entity, discrepancy, judgement, constitute, conserve |
| Loans | Lien, finance, fedloan, mortgage, disclosure, creditkarma, signature, student, jewel |
| Bankruptcy | Bankruptcy, validated, verified, convince, discharge, shellpoint, bsi |
| Forbearance | forbear, payment, repossession, accuracy, trace, overdrawn |
| Foreclosure | foreclosure, trust, income, qualify, ocwen, inspect |
| Unfair and Rude response | hung, chat, resurge, rude, garnish, talk, delink, whenever |

### F. RoBERTa

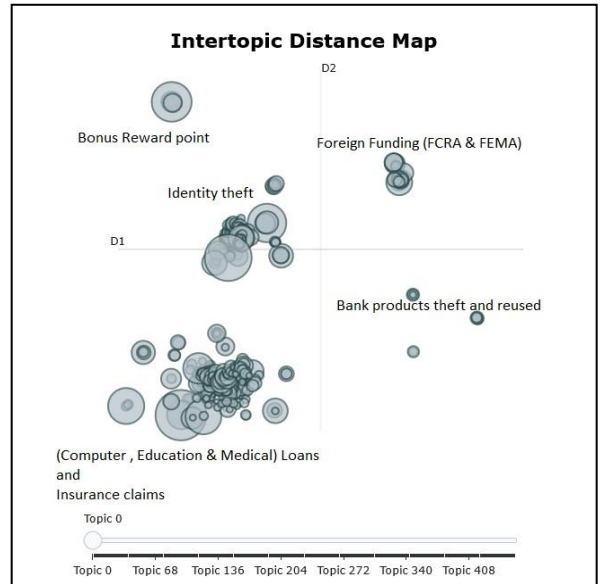

Figure 5. RoBERTa Topic modeling clusters

TABLE 7. ROBERTA TOPICS

| Topic Name | Major words in the cluster |
|---|---|
| Identification Theft | victim, modify, identify, theft, trust, employ, voice, fraudulent |
| Foreign Funding (FCRA & FEMA) | FCRA, thank, notify, shortage, fay, fema |
| Bonus Reward points | reward, membership, spend, bonus, promotion, advantage, consecutive |
| (Computer, Education & Medical) Loans or Insurance claims | rehabilitation, student, course, interest, mortgage, roadlaon, visa, suntrust, pma, college, costco, ira, atm |
| Bank products theft and reused | reinsert, grave, suffer, inquiry, reopen, grace, theft |

As per the inter-topic distance maps generated by various BERTopic based models, we can notice that BERT created twelve clusters (Figure 2), FinBERT produced ten clusters (Figure 3) & DistilBERT generated eleven clusters (Figure 4). However, RoBERTa generated only five clusters (Figure 5). After careful reviewing, we can observe that FinBERT efficiently generates diverse subjects, followed by DistilBERT and BERT. The topics generated by BERTopic using BERT (Table 4), FinBERT (Table 5), DistilBERT (Table 6), and RoBERTa (Table 7) were carefully evaluated by domain experts, and the corresponding topic name was assigned to the top ten words in each cluster.

TABLE 8. COMPARISON OF TOPICS CREATED BY DIFFERENT BERT BASED TOPIC MODELS

| Topics in common | BERT | FinBERT | DistilBERT | RoBERTa |
|---|---|---|---|---|
| 2 | eBanking | | eBanking or Digital Fraud | |
| | Delinquent Debt Collection | Collections | | |
| | Foreclosure | | Foreclosure | |
| | FICO and payday | | Payday Enquiries | |
| | Skymile Rewards | | | Bonus Reward Points |
| | | FCRA & FDCPA | | Foreign Funding (FCRA & FEMA) |
| 3 | Rude Customer Support | Glitches, violations related inquiries and their resolutions | Unfair and Rude response | |
| | Mortgage Loans | | Loans | (Computer, Education & Medical) Loans or Insurance Claims |
| | Delayed and Reraised | Customer Support (Knowledge & Communication) | Customer Care experience | |
| | Bankruptcy and Reinvestment | Delete discharge / Duplicate disclosures | Misinformation | |
| | Bankruptcy and Fraud | | Bankruptcy | Identification Theft |
| 4 | Product Delivery Related | Cashnet Issues | Regulatory Related | Bank Products Theft and Reused |
| 1 | Autopay | | | |
| | | | OTC Related Complaints | |
| | | | Forbearance | |
| | | Credit Reporting | | |
| | | Citi Bank Complaints | | |
| | | TransUnion Communications | | |
| | | FTB & FTC | | |

## G. Comparison of C_V and U_Mass

'C_V' and 'U_Mass' are the two evaluation parameters used to compare the efficacy of topic models. The C_V measure reflects topic similarity, whereas U_Mass assesses intra-word distance within subjects. A topic model is deemed efficient when the internal distance between the subjects is the shortest and the inter distance between the clusters is the greatest.

TABLE 9. PERFORMANCE METRICS COMPARISON OF DIFFERENT TOPIC MODELING TECHNIQUES

| Metrics/Algorithm | C_V | U_Mass |
|---|---|---|
| LSA | 0.2365 | -2.2675 |
| LDA | 0.3197 | -6.1207 |
| BERT | 0.3225 | -12.3169 |
| FinBERT | 0.3327 | -12.6526 |
| DistilBERT | 0.3249 | -11.2687 |
| RoBERTa | 0.3269 | -11.521 |

The C_V and U_Mass for each of the six topic modelling approaches utilized in this study are shown in the (Table 9). Compared to other topic modelling approaches, with a C_V score of 0.2365, LSA scored the lowest. While with a C_V score of 0.3327, FinBERT scored the highest. FinBERT as we expected, performed better with the dataset at hand since it was trained on financial data.

All of the topics created by BERT, DistilBERT, FinBERT, and RoBERTa are shown in (Table 8.) BERT, FinBERT, and DistilBERT share three topics: "Customer care experience, Unfair and Rude reaction and Misinformation" and two topics are shared by BERT, DistilBERT, and RoBERTa: "Loans and Bankruptcy". We can notice that BERT and DistilBERT generated similar subject clusters: "eBANKING, Customer service experience, Unfair and unpleasant answer, Misinformation, Loans, Bankruptcy, Payday Enquiry, and Regulatory linked". BERT could only identify one distinct topic, DistilBERT identified two distinct topics, while RoBERTa could not identify any distinct topics. FinBERT, on the other hand, was able to distinguish four separate topics because it was trained on finance domain data. The other BERTs had been trained on the Book Corpus and the Wikipedia English corpus.

## VI. CONCLUSION

The traditional techniques for topic modeling on financial text data were reviewed and compared. New and latest approaches like BERT, FinBERT, RoBERTa, and DistilBERT were tailored specifically for modeling the topics. Pre-trained models contain more accurate representations of words and sentences, and hence interpretation of topics is more straightforward in Bert-based models as stemming or lemmatization is not needed. BERTopics create easily interpretable topics with better C_V scores than traditional non-transformer-based NLP topic modelling techniques like LSA and LDA. Within the transformer-based, FinBERT performed better on our data as it is finetuned on financial documents, and it produces more distinct topics and better C_V among all BERT topics.

## References

[1] Agrawal, A., Fu W., Menzies T., What is wrong with topic modeling? And how to fix it using search-based software engineering, Inf. Softw. Technol. 98 (2018) 74–88.

[2] Bengio, Y., Ducharme, R., Vincent, P., Jauvin, C., Ca, J.U., Kandola, J., Hofmann, T., Poggio, T. and Shawe-Taylor, J., (2003) A Neural Probabilistic Language Model. Journal of Machine Learning Research.

[3] Bengio, Y., Schwenk, H., Senécal, J.-S., Morin, F. and Gauvain, J.-L., (2006) 6 Neural Probabilistic Language Models. [online] StudFuzz, Springer-Verlag. Available at: www.springerlink.com.

[4] Blei, D.M., Ng, A.Y. and Edu, J.B., (2003), Latent Dirichlet Allocation, J. Mach. Learn. Res. 3 (2003) (2003) 993–1022.

[5] Blei, D.M., Griffiths, T.L., Jordan, M.I. and Tenenbaum, J.B., (2003a) Hierarchical Topic Models and the Nested Chinese Restaurant Process.

[6] Blei, D.M. and Lafferty, J.D., Dynamic topic models, in: Proceedings of the 23rd International Conference on Machine Learning, Pittsburgh, PA, USA, 2006, pp. 113–120.

[7] Blei, D.M. and Lafferty, J.D., (2007) A Correlated Topic Model of Science. [online] 11, pp.17–35. Available at: http://imstat.org/aoas/supplements [Accessed 18 Mar. 2021].

[8] Christos H. Papadimitriou, Hisao Tamaki, Prabhakar Raghavan and Santosh Vempala., (2000) Latent Semantic Indexing: A Probabilistic Analysis. [online] Available at: https://reader.elsevier.com/reader/sd/pii/S0022000000917112?token=D8D7BD5339E981DA52463A733D90BDD3C84DB6261F644E6386994FAEBB26A291A56ADDBD65EC0C656B1FFABE5328B8F1 [Accessed 17 Mar. 2021].

[9] Dieng, A.B., Ruiz, F.J.R. and Blei, D.M., (2020) Topic Modeling in Embedding Spaces. [online] Available at: https://doi.org/10.1162/tacl.

[10] Daud, A., Li, J., Zhou, L. and Muhammad, F., (2010) Knowledge discovery through directed probabilistic topic models: A survey. Frontiers of Computer Science in China.

[11] Grootendorst, Maarten, (2020) BERTopic: Leveraging BERT and c-TF-IDF to create easily interpretable topics. Zenodo. Available at: https://doi.org/10.5281/zenodo.4381785

[12] Hofmann, T., (2001) Unsupervised Learning by Probabilistic Latent Semantic Analysis.

[13] Sharma, D., Kumar, B. and Chand, S., (2017) A Survey on Journey of Topic Modeling Techniques from SVD to Deep Learning. International Journal of Modern Education and Computer Science, [online] 97, pp.50–62. Available at: http://www.mecs-press.org/ijmecs/ijmecs-v9-n7/v9n7-6.html.

[14] Steyvers, M. and Griffiths, T., (2007) Probabilistic Topic Models.

[15] K. Vorontsov, A. Potapenko, Tutorial on probabilistic topic modeling: Additive regularization for stochastic matrix factorization, in: D.I. Ignatov, M.Y. Khachay, A. Panchenko, N. Konstantinova, R.E. Yavorsky (Eds.), Analysis of Images, Social Networks and Texts, Vol. 436, Springer International Publishing, Cham, 2014, pp. 29–46.

[16] Keerthi Kumar, H.M., Likhitha, S. and Harish, B.S., (2019) A Detailed Survey on Topic Modeling for Document and Short Text Data. Article in International Journal of Computer Applications, [online] 17839, pp.975–8887. Available at: https://www.researchgate.net/publication/335339697 [Accessed 18 Mar. 2021].

[17] Liangjie Hong and Brian D. Davison, (2010) Empirical Study of Topic Modeling in Twitter. Association for Computing Machinery.

[18] Rosen-Zvi, M., Griffiths, T., Steyvers, M. and Smyth, P., (2004) The Author-Topic Model for Authors and Documents. 102

[19] US Gov Agency, (2021) Consumer Complaint Database | Consumer Financial Protection Bureau. [online] Available at: https://www.consumerfinance.gov/data-research/consumer-complaints/ [Accessed 14 Jan. 2021].

[20] Xie, P. and Xing, E.P., (2013) Integrating Document Clustering and Topic Modeling.

[21] Nayak, D.K., Bolla, B.K. (2022). Efficient Deep Learning Methods for Sarcasm Detection of News Headlines. In: Chen, J.IZ., Wang, H., Du, KL., Suma, V. (eds) Machine Learning and Autonomous Systems. Smart Innovation, Systems and Technologies, vol 269. Springer, Singapore. https://doi.org/10.1007/978-981-16-7996-4_26

[22] Eranpurwala, F., Ramane, P., Bolla, B.K. (2022). Comparative Study of Marathi Text Classification Using Monolingual and Multilingual Embeddings. In: Woungang, I., Dhurandher, S.K., Pattanaik, K.K., Verma, A., Verma, P. (eds) Advanced Network Technologies and Intelligent Computing. ANTIC 2021. Communications in Computer and Information Science, vol 1534. Springer, Cham. https://doi.org/10.1007/978-3-030-96040-7_35